# Hate Speech detection in the Bengali language: A dataset and its baseline evaluation

Nauros Romim[1][0000−0002−2000−0898], Mosahed Ahmed[1][0000−0002−0010−397X], Hriteshwar Talukder[1,2][0000−0002−6391−6177], and Md Saiful Islam[1,2][0000−0001−9236−380X]

[1] Shahjalal University of Science and Technology, Kumargaon, Sylhet 3114, Bangladesh
{naurosromim,tainahmed96} @gmail.com
[2] {hriteshwar-eee,saiful-cse} @sust.edu

**Abstract.** Social media sites such as YouTube and Facebook have become an integral part of everyone's life and in the last few years, hate speech in the social media comment section has increased rapidly. Detection of hate speech on social media websites faces a variety of challenges including small imbalanced data sets, the finding of an appropriate model and also the choice of feature analysis method. Furthermore, this problem is more severe for the Bengali speaking community due to the lack of gold standard labelled datasets. This paper presents a new dataset of 30,000 user comments tagged by crowdsourcing and verified by expert. All the user comments collected from YouTube and Facebook comment section and to classified into seven categories: sports, entertainment, religion, politics, crime, celebrity, and TikTok & meme. A total of 50 annotators annotated each comment three times, and the majority vote was taken as the final annotation. Nevertheless, we have conducted baseline experiments and several deep learning models along with extensive pretrained Bengali word embedding such as Word2Vec, FastTest, and BengFastText on this dataset to facilitate future research opportunities. The experiment illustrated that although all the deep learning model performed well, SVM achieved the best result with 87.5% accuracy. Our core contribution is to make this benchmark dataset available and accessible to facilitate further research in the field of Bengali hate speech detection.

## 1 Introduction

Social media has become an essential part of every person's day to day life. It enables fast communication, easy access to sharing, and receiving ideas and views from worldwide. However, at the same time, this expression of freedom has lead to the continuous rise of hate speech and offensive language in social media.

Part of this problem has been created by the corporate social media model and the gap between documented community policy and the real-life implication of hate speech [4]. Not only that, but hate speech-language is also very diverse [17]. The language used in social media is often very different from traditional print media. It has various linguistic features. Thus the challenge in automatically detect hate speech is very hard [20].

Even though much work has been done in hate speech prevention in the English language, there is a significant lacking of resources regarding hate speech detection in Bengali social media. Nevertheless, problems like online abuse and especially online abuse towards females, are continuously on the rise in Bangladesh [19]. However, for language like Bengali, a low resource language, developing and deploying machine learning models to tackle real-life problems is very difficult, since there is a shortage of dataset and other tools for Bengali text classification [6]. So, the need for research on the nature and prevention of social media hate speech has never been higher.

This paper illustrates our attempt to improve this problem. Our dataset comprises 30,000 Bengali comments from YouTube and Facebook comment sections and has a 10,000 hate speech. We select comments from 7 different categories: *Sports, Entertainment, Crime, Religion, Politics, Celebrity, TikTok & meme*, making it diverse. We ran several deep learning models along with word embedding models such as Word2Vec, FastText, and pretrained BengFastText on our dataset to obtain benchmark results. Lastly, we analyzed our findings and explained challenges to detect hate speech.

## 2    Literature review

Much work has been done on detecting hate speech using deep learning models [7], [10]. There have also been efforts to increase the accuracy of predicting hate speech specifically by extracting unique semantic features of hate speech comments [22]. Researchers have also utilized fastText to build models that can be trained on billions of words in less than ten minutes and classify millions of sentences among hundreds of classes [14]. There have also been researches that indicate how the annotator's bias and worldview affect the performance of a dataset [21]. Right now, the state of the art research for hate speech detection reached the point where researches utilize the power of advanced architectures such as transfer learning. For example, in [9], researchers compared deep learning, transfer learning, and multitask learning architectures on the Arabic hate speech dataset. There is also research in identifying hate speech from a multilingual dataset using the pretrained state of the art models such as mBERT and xlm-RoBERTa [3].

Unfortunately, very few works have been done on hate speech detection in Bengali social media. The main challenge is the lack of sufficient data. To the best of our knowledge, many of the datasets were around 5000 corpora [5], [8] and [12]. There was a publicly available corpus containing around 10000 corpora, which were annotated into five different classes [2]. Nevertheless, one limitation the authors faced that they could only use sentences labeled as toxicity in their experiments since other labels were low in number. There was also another dataset of 2665 corpus that was translated from an English hate speech dataset [1]. Another research used a rule-based stemmer to obtain linguistic features [8]. Researches are coming out that uses deep learning models such as CNN, LSTM to obtain better results [2], [5], and [8]. One of the biggest challenges is that Bengali is a low resource language. Research has been done to create word embedding specifically for a low-resource language like Bengali called BengFastText [15]. This embedding was trained on 250 million Bengali word corpus. However, one thing is clear that there is a lack of dataset that is both large and diverse. This paper tries to tackle the problem by presenting a large dataset and having comments from seven different categories, making it diverse. To our knowledge, this is the first dataset in Bengali social media hate speech with such a large and diverse dataset.

## 3  Dataset

### 3.1  Data extraction

Our primary goal while creating the dataset was to create a dataset with different varieties of data. For this reason, comments on seven different categories: *sports, entertainment, crime, religion, politics, celebrity and meme, TikTok & miscellaneous* from YouTube and Facebook were extracted.

We extracted comments from the Facebook public page, Dr. Muhammad Zafar Iqbal. He is a prominent science fiction author, physicist, academic, and activist of Bangladesh. These comments belonged to the *celebrity* category. Nevertheless, due to Facebook's restriction on its graph API, we had to focus on YouTube as the primary source of data.

From YouTube, we looked for the most scandalous and controversial topics of Bangladesh between the year 2017-20. We reasoned that since these were controversial, videos were made more frequently, and people participated more in the comment section, and the comments might contain more hate speech. We searched YouTube for videos on keywords related to these controversial events. For example, we searched for renowned singer and actor couple *Mithila-Tahsan divorce*, i.e., the Mithila controversy of 2020. Then we selected videos with at least 700k views and extracted comments from them. In this way, we extracted comments videos on controversial topics covering five categories: *sports, entertainment, crime, religion, and politics.* Finally, we searched for videos that are memes, TikTok, and other keywords that might contain hate speech in their comment section. This is our seventh category.

We extracted all the comments using open-source software called FacePager[3]. After extracting, we labeled each comment with a number defining which category it belonged. We also labeled the comments with the additional number to define its *keyword*. For this paper, the *keyword* means the controversial event that falls under a *category*. For example, *mithila* is a keyword that belongs to the *entertainment* category. We labeled every comment with its corresponding category and keyword for future research.

---

[3] https://github.com/strohne/Facepager

After extracting the comments, we manually checked the whole dataset and removed all the comments made entirely of English or other non-Bengali languages, emoji, punctuation, and numerical values. However, we kept a comment if it primarily consists of Bengali words with some emoji, number, and non-Bengali words mixed within it. We deleted non-Bengali comments because our research focuses on Bengali hate speech comments, so non-Bengali comments are out of our research focus. However, we kept impure Bengali comments because people do not make pure Bengali comments on social media. So emoji, number, punctuation, and English words are a feature of social media comment. Thus we believe our dataset can prove to be very rich for future research purposes. In the end, we collected a total of 30,000 comments. In a nutshell, our dataset has 30k comments that are mostly Bengali sentences with some emoji, punctuation, number, and English alphabet mixed in it.

## 3.2   Dataset annotation

Hate speech is a subjective matter. So it is quite difficult to define what makes a comment hate speech. Thus we have come up with some rigid rules. We have based these rules on the community standard of Facebook[4] and YouTube[5]. Bellow, we have listed some criteria with necessary examples:

– Hate speech is a sentence that dehumanizes one or multiple persons or a community. Dehumanizing can be done by comparing the person or community to an insect, object, or a criminal. It can also be done by targeting a person based on their race, gender, physical and mental disability.
– A sentence might contain slang or inappropriate language. But unless that slang dehumanizes a person or community, we did not consider it to be hate speech. For example:

কি বালের মুভি

Here the slang word is not used to dehumanized any person. So it is not hate speech.
– If a comment does not dehumanize a person rather directly supports another idea that clearly dehumanizes a person or a community, This is considered hate speech. For example:

বোরকা পরে না, ধর্ষিত তো হবেই।

this sentence supports the dehumanizing act of women, so we labeled it as hate speech.

---

- If additional context is needed to understand that a comment is a hate speech, we did no consider it to be one. For example, consider this sentence:

গর্ত যোদ্ধা

It is a comment taken from Dr. Muhammad Zafar Iqbal's Facebook page. This comment refers to a particular jab the haters of Dr. Zafar Iqbal constantly use to attack him in social media. But unless no one mentions an annotator that this comment belongs to this particular Facebook page, that person will have no way of knowing this is actually a hate speech. Thus these types of comments will be labeled as not hate speech.

- It does not matter if the stand that a hate speech comment takes is right or wrong. Because what is right or wrong is subjective. So if a sentence, without any outside context, dehumanizes a person or community, we considered that to be hate speech.

We worked with 50 annotators to annotate the entire dataset. We instructed all annotators to follow our guidelines mentioned above. The annotators are all undergraduate students of Shahjalal University of Science and Technology. Thus the annotators have an excellent understanding of popular social media trends and seen how hate speech propagates in social media. All comments were annotated three times, and we took the majority decision as the final annotation.

After annotation, we wanted to check the validity of the annotator's annotation. For this reason, we randomly sampled 300 comments from every category. Then we manually checked each comment's majority decision. Since we, as the authors of this paper, were the ones that set the guideline for defining hate speech and did not participate in the annotation procedure, our checking was a neutral evaluation of the annotator's performance. After our evaluation, we found that our dataset annotation is 91.05% correct.

### 3.3 Dataset statistics

Our dataset has a total of 30,000 comments, where there are a total of 10,000 hate speech comments, as evident from table-2. So it is clear that our dataset is heavily biased towards not hate speech. If we look closely at each category in figure-1, it becomes even more apparent. In all categories, not hate speech comments dominates. But particularly in *celebrity* and *politics*, the number of hate speech is very low, even compared to hate speech in other categories. During data collection, we have observed that there were many hate speech in the *celebrity* section i.e. in Dr. Muhammad Zafar Iqbal's Facebook page, but they were outside context. As we have discussed before in section 3.2, we have only considered texts without context while labeling it as hate speech. So many comments were labeled as not hate speech. For the category *politics*, we have observed that people tend to not attack any person or group directly. Rather they tend to add their own take on the current political environment. So the number of direct attacks is less in the *politics* category.

When we look at the mean text length in table-3, we can find a couple of interesting observations. First, we can see that meme comments are very short in length. This makes sense as when a person is posting a comment in a meme video, and he is likely to express his state of mind, requiring a shorter amount sentences. But the opposite is true for the *celebrity* category. This has the longest average text length. This is large because when people comment on Dr. Zafar Iqbal's Facebook page, they add a lot of their own opinion and analysis, no matter the comment is hate speech or not. This shows how unique the comment section of an individual celebrity page can be. Lastly, we see that average hate speech tends to be shorter than not hate speech.

In the table, 4 we have compared all state of the art datasets. The table we have included the total number of the dataset and the number of classes the datasets were annotated. As you can see, there are some datasets that have multiple classes. In this paper, we focused on the total number of the dataset and extracted comments from different categories so that we can ensure linguistic variation.

**Table 1.** Sample dataset

| Sentence | Hate speech | Category | Keyword |
|---|---|---|---|
| মহিলার বিচার চাই | 0 | 3 | 1 |
| হারপো তোর কত বড় সাহস? | 1 | 1 | 6 |
| আমরা পাশে আছি ব্রাদার। | 0 | 2 | 12 |
| এই গাধার বাচ্চা, কি বলিস তুই | 1 | 3 | 3 |
| হে আল্লাহ, তার সুস্থতা দা কর | 0 | 6 | 1 |

**Table 2.** Hate speech comments per category

| Hate speech | Not hate speech | Total |
|---|---|---|
| 10,000 | 20,000 | 30,000 |

## 4    Experiment

### 4.1    Preprocess

Our 30k dataset had raw Bengali comments with emoji, punctuation, and English alphabet mixed in it. We removed all emoji, punctuation, numerical values, non-Bengali alphabet, and symbols from all comments for baseline evaluation. After that, we created a train and test set of 24,000 and 6,000 comments, respectively. Now the dataset is ready for evaluation.

**Table 3.** Mean text length of the dataset

| Category | Mean text length |
|---|---|
| Sports | 75 |
| Entertainment | 65.9 |
| Crime | 87.4 |
| Religion | 71.4 |
| Politics | 72.5 |
| Celebrity | 134.5 |
| Meme, TikTok and others | 56.2 |
| Hate speech | 69.59 |
| Not hate speech | 84.39 |

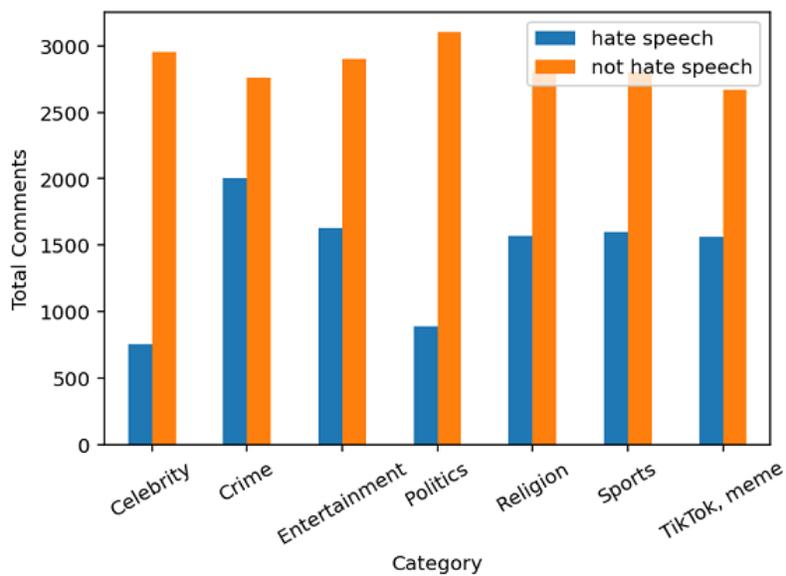

**Fig. 1.** Distribution diagram of data in each category

**Table 4.** A comparison of all state of the art datasets on Bengali hate speech

| Paper | Total data | Number of classes |
|---|---|---|
| Hateful speech detection in public facebook pages for the bengali language [12] | 5,126 | 06 |
| Toxicity Detection on Bengali Social Media Comments using Supervised Models [2] | 10,219 | 05 |
| A Deep Learning Approach to Detect Abusive Bengali Text [8] | 4,700 | 07 |
| Threat and Abusive Language Detection on Social Media in Bengali Language [5] | 5,644 | 07 |
| Detecting Abusive Comments in Discussion Threads Using Naïve Bayes [1] | 2,665 | 07 |
| **Hate Speech detection in the Bengali language: A dataset and its baseline evaluation** | **30,000** | **02** |

### 4.2 Word embedding

We have used three word embedding models. They are: Word2Vec [16], FastText [13] and BengFast [18]. To create a Word2Vec model, we used gensim[6] module to train on the 30k dataset. We have used CBoW method for building the Word2Vec model. For FastText, we also used the 30k dataset to create the embedding and used the skip-gram method. The embedding dimension for both models was set to 300. Lastly, BengFastText is the largest pretrained Bengali word embedding based on FastText. But BengFastText was not trained on any YouTube data. So we wanted to see how it performs on YouTube comments.

### 4.3 Models

**Support Vector Machine (SVM)** We used the Support Vector Machine [11] to determine baseline evaluation. We used the linear kernel and kept all other parameters to its default value.

**Long Short Term Memory (LSTM)** For our experiment, we used 100 LSTM layers, set both dropout and recurrent dropout rate to 0.2, and used 'adam' as the optimizer.

**Bi-directional Long Short Term Memory (Bi-LSTM)** In this case, we used 64 Bi-LSTM layers, with a dropout rate set to 0.2 and optimizer as 'adam.'

### 4.4 Experimental setting

We kept 80% of the dataset as a train set and 20% as a test and trained every word embedding with every deep learning algorithm on the train set. For every case, we kept all parameters standard. Epoch and batch size was set to 5 and 64, respectively. Then we tested all the trained models on the test dataset and measured the accuracy and F-1 score. Bellow are all types of models we tested on our dataset.

---

[6] *https : //radimrehurek.com/gensim/models/word2vec.html*

- Baseline evaluation: Support Vector Machine (SVM)
- FastText Embedding with LSTM
- FastText Embedding with Bi-LSTM
- BengFastText Embedding with LSTM
- BengFastText Embedding with Bi-LSTM
- Word2Vec Embedding with LSTM
- Word2Vec Embedding with Bi-LSTM

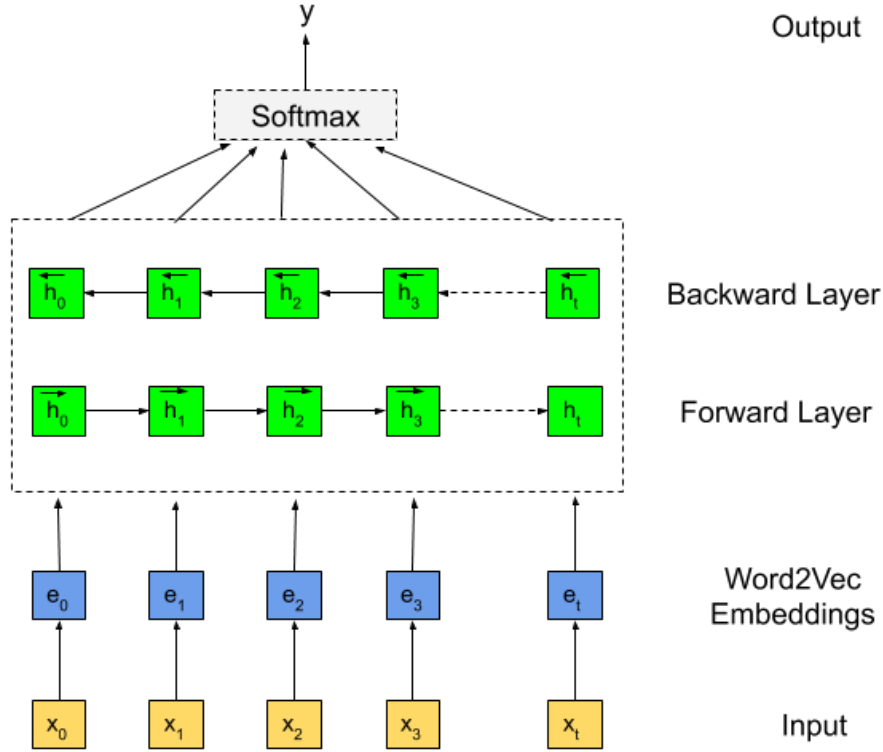

**Fig. 2.** Deep learning architecture with Word2Vec and Bi-LSTM

### 4.5    Result

We can observe from the table 5 that all the models achieved good accuracy. SVM achieved the overall best result with accuracy and an F-1 score of 87.5% and 0.911, respectively. But BengFastText with LSTM and Bi-LSTM had relatively the worst accuracy and F-1 score. Their low F-1 score indicates that deep learning models with BengFastText embedding were overfitted the most. BengFastText is not trained on any YouTube data [18]. But our dataset has a huge amount of YouTube comment. This might be a reason for its drop on performance.

Then we looked at the performance of Word2Vec and FastText embedding. We can see that FastText performed better in terms of accuracy and had a lower F-1 score than Word2Vec. Word2Vec was more overfitted than FastText. FastText has one distinct advantage over Word2Vec: it learns from the words of a corpus and it's substrings. Thus FastText can tell 'love' and 'beloved' are similar words [13]. This might be a reason as to why FastText outperformed Word2Vec.

**Table 5.** Result of all models

| Model name | Accuracy | F-1 Score |
| --- | --- | --- |
| SVM | **87.5** | **0.911** |
| Word2Vec + LSTM | 83.85 | 0.89 |
| Word2Vec + Bi-LSTM | 81.52 | 0.86 |
| FastText + LSTM | 84.3 | 0.89 |
| FastText + Bi-LSTM | 86.55 | 0.901 |
| BengFastText + LSTM | 81 | 0.859 |
| BengFastText + Bi-LSTM | 80.44 | 0.857 |

### 4.6    Error analysis

We manually cross-checked all labels of the test set with the prediction of the SVM model. We looked at the false negative and false positive cases and wanted to find which types of sentences the model failed to predict accurately. We found that some of the labels were actually wrong, and the model actually predicted accurately. Nevertheless, there were some unusual cases. For example:

ওরা চাইবো কেন তোর কাছে তুই বিসিবি প্রেসিডেন্ট হয়ে কি বাল ফালাস তোর তো খেয়াল রাখা দরকার ছিল ক্রিকেটারদের ফিসিলিস নিয়ে

This is not a hate speech, but the model predicted it to be hate speech. There are several other similar examples. The reason here is that this sentence contains aggressive words that are normally used in hate speech, but in this case, it was not used to dehumanize another person. However, the model failed to understand that. This type of mistake was common in false-negative cases. It demonstrates that words in the Bengali language can be used in a complicated context, and it is a tremendous challenge for machine learning models to actually understand the proper context of a word used in a sentence.

## 5   Conclusion

Hate speech comment in social media is a pervasive problem. So a lot more research is urgent to combat this issue and ensure a better online environment. One of the biggest obstacles towards detecting hate speech through state-of-the-art deep learning models is the lack of a large and diverse dataset. In this paper, we created a large dataset of 30,000 comments, a 10,000 hate speech. Our dataset has comments from 7 different categories making the dataset diverse and abundant. We showed that hate speech comments tend to be shorter in length and word count than not hate speech comments. Finally, we ran several deep learning models with word embedding on our dataset. It showed that when the training dataset is highly imbalanced, the models become overfitted and biased towards not hate speech. Thus even though overall accuracy is very high, the models can not predict hate speech well.

However, we believe this is scratching just the surface of solving this widespread problem. One of the biggest obstacles is that there is no proper word embedding for the Bengali language used in social media. There is some word embedding created from newspaper and blog article corpus. Nevertheless, the language used is vastly different from social media language. The main reason is that unlike traditional print media, there is no one to check for grammatical and spelling mistakes. Thus there are lots of misspelling, grammatical errors, cryptic meme language, emoji, etc. In fact, in our dataset, we found the same word having multiple types of spelling. For example:

<div align="center">জাব, যাব, যাবো, জাবও</div>

Here, they all are the same word. A human brain can understand that these words are the same, but they are different from a deep learning model. Another difference is the use of emoji to convey meaning. Often people will express a specific emotion with the only emoji. Emoji is a recent phenomenon that is absent in blog posts or newspaper articles, or books. Currently, there is no dataset or pretrained model that classifies the sentiment of emoji used in social media.

One of the critical challenges to accurate hate speech detection is to create models that can extract necessary information from an unbalanced dataset to predict the minority class with reasonable accuracy. Our experiment demonstrated that standard deep learning models are not sufficient for this task. Advanced models like mBERT and XLM-RoBERTa can be of great use in this regard as they are trained on a large multilingual dataset and use attention mechanisms. Embedding models based on extensive and diverse Social Media comment datasets can also be of great help.

## 6    Acknowledgement

This work would not have been possible without the kind support from SUST NLP Research Group and SUST Research Center. We would also like to express our hearfealt gratitude to all the annotators and volunteers who made the journey possible.